\documentclass{article}
\usepackage[utf8]{inputenc}
\usepackage{arxiv}
\usepackage{amsmath}
\usepackage{enumerate}
\usepackage{subfigure}
\usepackage{graphicx}
\usepackage{booktabs}
\usepackage{natbib}

\graphicspath{{.}}

\usepackage[inline,shortlabels]{enumitem}

\usepackage[nameinlink]{cleveref}
\crefname{figure}{Fig.}{Figs.}
\Crefname{figure}{Figure}{Figures}
\crefname{equation}{Eq.}{Eqs.}
\Crefname{equation}{Equation}{Equations}
\crefformat{section}{#2§#1#3}
\crefname{section}{§}{§§}
\Crefname{section}{Section}{Sections}
\crefname{table}{Table}{Tables}
\crefname{appendix}{Appendix}{Appendices}

\newcommand{\eg}{e.g.,\ }

\newcommand{\ie}{i.e.,\ }

\title{Transfer Learning in Transformer-Based Demand Forecasting For Home Energy Management System}

\author{%
Gargya Gokhale~\thanks{Presented at BALANCES Workshop at the 10th ACM International Conference on Systems for Energy-Efficient Buildings, Cities, and Transportation (BuildSys '23), November 15--16, 2023, Istanbul, Turkey. Publication rights licensed to ACM. \\ DOI: 10.1145/3600100.3626635.}\\
IDLab, Ghent University--imec\\
Gent, Belgium \\
\texttt{gargya.gokhale@ugent.be} \\
\And
Jonas Van Gompel\\
IDLab, Ghent University--imec\\
Gent, Belgium\\
\And
Bert Claessens\\
IDLab, Ghent University--imec\\
beebop.ai \\
\And
Chris Develder\\
IDLab, Ghent University--imec\\
Gent, Belgium\\
}

\begin{document}

\maketitle

\begin{abstract}
Increasingly, homeowners opt for  photovoltaic (PV) systems and/or battery storage to minimize their energy bills and maximize renewable energy usage. This has spurred the development of advanced control algorithms that maximally achieve those goals.
However, a common challenge faced while developing such controllers is the unavailability of accurate forecasts of household power consumption, especially for shorter time resolutions (15 minutes) and  in a data-efficient manner. In this paper, we analyze how transfer learning can help by exploiting data from multiple households to improve a single house's load forecasting. Specifically, we train an advanced forecasting model (a temporal fusion transformer) using data from multiple different households, and then finetune this global model on a new household with limited data (\ie only a few days). The obtained models are used for forecasting power consumption of the household for the next 24 hours~(day-ahead) at a time resolution of 15 minutes, with the intention of using these forecasts in advanced controllers such as Model Predictive Control. We show the benefit of this transfer learning setup versus solely using the individual new household's data, both in terms of
\begin{enumerate*}[(i)]
    \item forecasting accuracy ($\sim$15\% MAE reduction) and
    \item control performance ($\sim$2\% energy cost reduction),
\end{enumerate*}
using real-world household data.
\end{abstract}


%===================================================================
\section{Introduction}
\label{sec:intro}
%===================================================================
% - High res, intermittent res, lack of flex/need for flex
% - HEMS -- batteries-- cost opt -- 
% - Advanced control-- MPC as dominant
% - Needs forecasts for demand
% - Significant work in forecasting-- hot topic-- mostly hourly for individual buildings
% - Quarter hour is the future -- elia ccmd etc.
% - Quarter hour is tough -- too much uncertainty 
% - Household data is difficult to get as well
% - What if we pre-train global models on different datasets and then fine tune for specific houses
% - Use transformers to see useful patterns and train on loads on data
% - Fine tune and show that this is much better than locally trained models

% general intro of modern grid challenges & advent of prosumers
The need for a clean and sustainable energy sector has led to significant changes in the modern power grid, including increased integration of renewable energy sources, use of advanced sensors and monitoring devices and growth in electrification~\cite{IEA_2023}.
Such changes have been pivotal in the rise of prosumers (\ie active energy consumers that produce and consume energy)~\cite{EEA_prosumer}.
For our work, we focus on households and residential prosumers operating with a financial objective of reducing energy bills.
In most cases, such residential prosumers rely on PV systems for producing electricity and consuming it instantaneously, with the option of injecting the excess to the power grid.
%However, %with
% transition to your particular HEMS problem
Driven by increased volatility in the energy markets~\cite{price-volatility}, %more and more residential prosumers are shifting towards 
we note an increasing shift to storage-based PV systems and more elaborate home energy management systems~(HEMS).
Supported by increased adoption of sensors, smart meters and other IoT devices, these HEMS make use of advanced control algorithms to identify optimum control strategies for individual prosumers.
% We focus on households and residential prosumers with a solar photovoltaic~(PV) system for producing electricity and a home battery for energy storage. These prosumers are driven by the financial objective of reducing energy bills and with increased price volatility, such prosumers are shifting towards more elaborate home energy management systems~(HEMS)
% As presented in~\cite{hems_review}, HEMS have been a major field of research, with prior works studying different control strategies for home energy management systems. 
Model Predictive Control~(MPC) has been a dominant control strategy for HEMS, with works such as~\cite{stochastic_mpc1, milp_mpc_real} exploring its applications in diverse settings.
% MPC is an advanced control strategy that relies on a model of the system and an optimizer~\cite{mpc_basics}.
% The optimizer uses the system model to anticipate the future behavior of the system and calculate an optimum control strategy to achieve the required objective.
In the HEMS context, an MPC entails using a battery model along with forecasts for PV production and household demand to model the household and then using standard optimization algorithms to obtain suitable control policies~\cite{mpc_basics}. However, obtaining accurate forecasts for individual household-level demand and PV production has been a major challenge.

Demand and PV production forecasting is an established research domain with works such as~\cite{load_forecasting_review, review_data_driven_consumption} presenting an overview of forecasting techniques specific to household-level demand and PV.
% As discussed in~\cite{load_forecasting_review}, p
Prior works have focused on %a plethora of forecasting
a range of techniques such as ARIMA-based models~\cite{arima_1, arima_2}, hybrid models~\cite{hybrid_forecast} as well as advanced deep learning methods~\cite{half_hourly_rnn, lstm_evo_ghop_load_forecast}, including recently introduced transformer architectures~\cite{TFT_load_forecasting_2023}.
While these recent deep learning-based methods show significant improvements in forecasting accuracy, a common problem associated with these methods is their susceptibility to overfitting and the need for large amounts of data to avoid it~\cite{energy_model_challenges}.

Works such as~\cite{tl_graph, tl_cnn_gru} have discussed the use of transfer learning to address this problem. However, these are limited to either hourly time resolutions or work with aggregated forecasts instead of modeling individual households. Only a few previous works such as~\cite{lstm_evo_ghop_load_forecast, half_hour_gru} have utilized a time resolution of less than an hour for their forecasting problems.
Forecasting electricity demand of an individual household on a quarter hour basis is an extremely challenging problem, primarily due to the significant influence of user-behavior, which can vary wildly and is difficult to model.
% and increased uncertainty corresponding to it. 
Since future power consumption is an important input to optimize HEMS decisions, improving the quality of such forecasts, especially for a quarter hour frequency, can boost the performance of the HEMS, allowing these prosumers to participate more effectively in future energy and consumer-centric markets~\cite{elia_ccmd}.
% This represents a significant gap in existing literature and we present our study to address this gap. 

The main goal of our work is to investigate how transfer learning methods can help utilize data from multiple different households to improve load forecasts of individual households and develop better control strategies for them. The transfer learning method presented in this work uses data from multiple households to train a global model. This global model can then be finetuned on a new household using only a few days worth of data to obtain a high performing forecasting model for that household. We validate this transfer learning methodology on real-world data obtained from 30 different households and by using the state-of-the-art Temporal Fusion Transformer~(TFT)~\cite{TFT} as the forecasting model. Through our simulations, we analyze the day-ahead, quarter hour resolution forecasting performance of our finetuned TFT models as well as the control strategies obtained using these forecasts. Our main contributions can be summarized as:
\begin{enumerate}
    \item We propose a transfer learning-based forecasting method using Temporal Fusion Transformers for day-ahead forecasting of individual household-level demand with a quarter hour frequency.
    \item We show that the fine-tuned models require less training data, can generalize to unseen households, and can outperform locally trained TFT models. 
    \item Using a simple MPC, we show that such fine-tuned models are effective in obtaining good control policies in home energy management systems. 
\end{enumerate}

%%

%===================================================================
\section{Problem Formulation}
\label{sec:problem_formulation}
%===================================================================
% - Describe the control problem and where forecasts are used
% - Describe data and show plots
% - Describe forecasting problem

As discussed in~\cref{sec:intro}, we focus on households and residential prosumers. We consider a single household with a PV system, a small residential battery~(5kW, 10kWh), and a dynamic energy tariff. The objective is to develop an energy management system that can minimize the electricity cost for this household by effectively utilizing the battery based on expected demand and PV generation of the household. 
To develop this energy management system, we formulate a simple MPC-based controller that works with a linear model of the battery and forecasted demand and PV generation profiles. This MPC is designed for a quarter hour control frequency and requires day-ahead forecasts of demand and PV generation profiles. 

\subsection{MPC Formulation}
The objective of the MPC-based energy management system is to minimize the cost of energy consumed by the household. We model the battery using a simple linear model with constant round trip efficiency~($\eta=90\%$). We assume a dynamic energy tariff with time-varying prices for consuming energy~($\lambda^{\text{con}}_{t}$) and injecting energy~($\lambda^{\text{inj}}_{t}$) to the grid.\footnote{We assume that price obtained for injecting energy into the grid is 40\% of the price paid for energy consumption. This number can vary depending on the energy contract.} Relying on forecasted values of PV generation~($P^{\text{pv}}_{t}$) and demand~($P^{\text{con}}_{t}$), the MPC must choose battery actions~($u_{t}$) at each time step~($t$) to minimize the cost of energy consumed over a horizon~($T$). For our problem, $\Delta t = 15 \text{mins}$. This optimization problem is presented in~\cref{eq:mpc}. Here, $E_{t}$ refers to the energy state of battery at step $t$, while $E^{\text{max}}$, $u^{\text{min}}$ and $u^{\text{max}}$ are the energy and power constraints of the battery. $P^{\text{G}}_{t}$ is the power at the meter and we denote power consumed with positive values. %Similarly for the battery, we denote charging power with positive values.
\begin{equation}
    \begin{split}
        \min_{u_{1}, \ldots u_{T}} &\sum_{t=1}^{T} c_{t} \\
        \text{s.t.:} \ c_{t} &= \begin{cases}
                                \lambda^{\text{con}}_{t} P^{\text{G}}_{t} \Delta t &: P^{\text{G}}_{t} \geq 0      \\
                                \lambda^{\text{inj}}_{t} P^{\text{G}}_{t} \Delta t &: P^{\text{G}}_{t} < 0\\
                                \end{cases} \quad \forall t \\
                        P^{\text{G}}_{t} &= P^{\text{con}}_{t} + P^{\text{pv}}_{t} + u_{t} \qquad \qquad \ \forall t \\
                        E_{t+1} &= \begin{cases}
                                E_{t} + \eta u_{t} \Delta t &: u_{t} \geq 0      \\
                                E_{t} + \frac{1}{\eta} u_{t} \Delta t &: u_{t} < 0\\
                                \end{cases} \quad \forall t \\
                              0 \leq E_{t} &\leq E^{\text{max}}; \
                              u^{\text{min}} \leq u_{t} \leq u^{\text{max}}  \quad \forall t
    \end{split}
\label{eq:mpc}
\end{equation}
Note that, this MPC is formulated as a simple, linear MPC, utilizing a linear model of the battery along with forecasted values of energy consumption and PV generation~(\ie the forecastors are not used in the optimization).
% As common with any MPC~\cite{mpc_basics}, this optimization problem is solved for each time step $t$ to obtain the control action $u_{t} = u_{1}$. To reduce the computational intensity of the problem, we implement a receding horizon approach, where the horizon $T$ is reduced after every time step, starting from $T=96$ for 00:00 to $T=1$ for 23:45. 

\subsection{Demand Forecasting}
The MPC problem formulated in~\cref{eq:mpc}, requires forecasted values for PV generation and electricity demand of the household. For this work, we focus on forecasting only the demand~($P^{\text{con}}_{t}$) and assume exact values for PV production. However, our methods can be extended to PV production forecasting as well. With the intention of using these forecasts for control applications, we model this electrical demand forecasting problem as a univariate stochastic forecasting problem.

% We model this electrical demand forecasting problem as a univariate forecasting problem. With the intention of using these forecasts further for control applications, the forecasting problem is modeled as a stochastic forecasting problem. 
% This leads to a forecasting model~(discussed in~\cref{sec:methodology}) that provides probabilistic forecasts over the horizon $T$. 

%===================================================================
\section{Methodology}
\label{sec:methodology}
%===================================================================
% - Explain transformer/ TFT
% - Explain global and fine tuning
% - Explain benchmarks
% - Explain MPC(?) formulation and solving
This section describes our transfer learning-based forecasting approach and provides details related to the forecasting model (Temporal Fusion Transformer) and the experimental setup used for our simulations.
% We present a forecasting approach that utilizes transfer learning and Temporal fusion transformer~(TFT). In this section, we describe these components and detail our experimental setup. 
\subsection{Temporal Fusion Transformer~(TFT)}
Building upon the success of the attention mechanism and the transformer architecture in natural language processing domains, the TFT architecture was proposed for multi-horizon time series forecasting~\cite{TFT}. In addition to the self-attention and cross-attention layers used in transformers, TFT uses specialized components, such as variable selection networks and gated connections, to effectively encode temporal relationships and obtain an interpretable and accurate forecasting model. These specialized components and the transformer architecture enable TFT to operate on long input series and efficiently capture long-term trends and dependencies in such series~\cite{TFT}. 

For our work, we use the TFT implementation from Darts~\cite{darts}. Each input consists of a series of past 672 quarter hours (7 days) of electrical demand for the house, along with temporal features such as hour of day, day of week, etc.\footnote{While works such as~\cite{TFT_load_forecasting_2023} recommend using other covariates such as outside air temperature, unavailability of such data led to this design choice.} The model output is the predictions for the next day’s demand (96 steps). We train our TFT models as stochastic predictors using a quantile regression loss function~\cite{qunatile_regression}. This leads to trained models that are able to predict when peaks in demand are likely, whereas deterministic forecasters trained with, \eg mean squared error loss are not able to capture this uncertainty. Further, our quantile forecasts can also be adapted to work with stochastic MPCs in the future. 
\subsection{Transfer Learning}
Transfer learning is used to improve a model from one domain by transferring information from a related domain~\cite{TL_survey}. In deep learning, transfer learning has been applied to leverage large amounts of data from different sources to pre-train a large model followed by finetuning this model on limited data from the target domain. We follow a similar approach to first train a global TFT model, followed by finetuning this global TFT on individual household’s data. We now describe the training and fine-tuning steps for our work. 
\subsubsection{Global Model}
The main idea behind a global model is to use a large set of data to learn good representations related to the commonly occurring patterns in the data. For our case, a global TFT model was trained using 15 months of data from 25 different buildings, amounting to approximately 1M data points. The global TFT model was trained using a quantile regression loss function and had a forecast horizon of 96 steps~(24 hours). About 15\% of the data was used as validation set for linearly decaying the learning rate and early stopping to avoid overfitting. 
\subsubsection{Finetuned Models}
The global model obtained from the previous step forms the base forecasting model that is to be finetuned for individual households. The finetuning step used a similar training loop as the global model. However, the learning rates and number of epochs in the training loop were significantly reduced. This ensures that during the finetuning phase, the changes to the weights of the TFT are limited and the model does not overfit on the small dataset corresponding to a single household. 

\subsection{Experiment Setup}
The transfer learning approach presented above was implemented on the data obtained as a part of the BD4NRG research project. This dataset corresponds to 30 real-world households for a period of close to 18 months and contains quarterly measurements of power consumption. The data was preprocessed using standard methods that included filtering null values, aligning time series, and removing outliers. Following this, data from 5 households was held out as test set. The remaining 25 households were used to pretrain the global model. The data from these 25 households was first scaled using a min-max scaler and then split into training and validation sets (85\% - 15\%). 
% These training and validation sets were used to pretrain the global TFT model. 
For finetuning, we used data from households in the held-out test set. Finetuning was performed using training data of different sizes~(from 14 days to 42 days worth of data) and 1 week worth of validation data. This was followed by testing on 6 weeks of unseen test data. To ensure a common test score across all training sizes, the test set was fixed, and training days corresponded to the $n$ days preceding this test set. % 1st Oct to 15th Nov for test
The results presented in \cref{sec:results} correspond to this test data.

%===================================================================
\section{Results and Discussion}
\label{sec:results}
%===================================================================
% - Show results for individual forecast
% - Show results for the 5 house performance for different data sizes
% - Show MPC results using local and finetuned forecast
The main idea behind the proposed transfer learning-based forecasting method is to obtain data-efficient forecasts on an individual household-level that can then be used for developing home energy management controllers. To investigate this idea, we first studied the forecasting performance of the finetuned TFT models followed by an evaluation of the control policy obtained when using forecasts from these models. In both cases, the finetuned TFT models were compared with ``local'' TFT models, \ie TFT models initialized with random weights and trained solely on data obtained from a single household. This comparison allows us to examine the added value of transfer learning for forecasting performance as well as control performance. 

\subsection{Forecasting Performance}
\begin{figure}[t]
    \centering
    \includegraphics[width=.95\columnwidth]{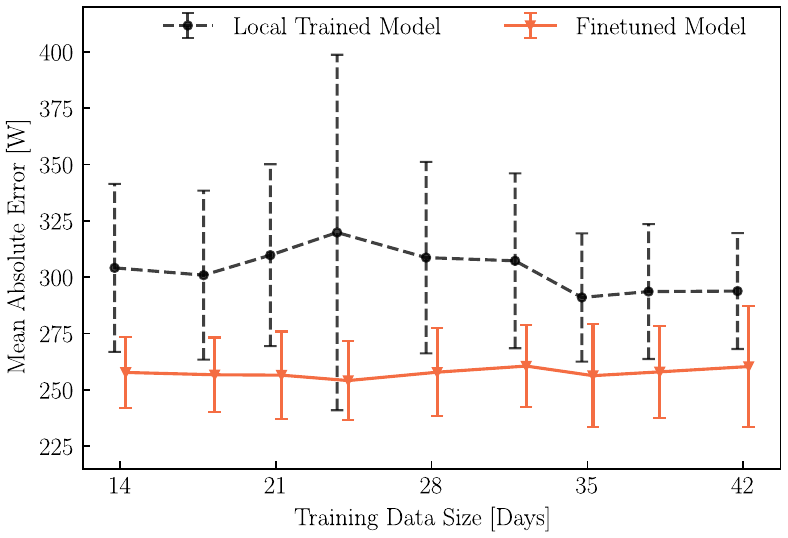}
    \caption{Forecasting Performance of Finetuned TFT model compared with TFT models trained using only local data. The points represent average MAE values over the 5-test households and the error bars represent the standard deviation.}
    \label{fig:forecast_perf}
\end{figure}

We tested the forecasting performance of our finetuned TFT models by providing the model with appropriate inputs at the beginning of each test day and obtaining their forecasts for the next 24 hours~(to follow the MPC formulated in \cref{sec:problem_formulation}).
Each input contained past observations only~(\ie forecasts of the model were never used as inputs) and the process was repeated over the entire test set~(6 weeks). The forecasts were compared using Mean Absolute error~(MAE). \Cref{fig:forecast_perf} presents the comparison of forecasting performance between our finetuned TFT models and TFT models trained only on local data. The markers depict the mean MAE over the 5-test households and the error bars indicate the standard deviation. It is evident from the figure that the finetuned models are performing significantly better~($\sim 15 \%$) than the local TFT models.

% Note that, increasing training data doesn't have a big impact on local TFT models. A possible explanation is that more days preceding the test set are given, 
% Note that, increasing training data doesn't have a big impact on local TFT models. A possible explanation can be the way training and test data set were chosen. As discussed in~\cref{sec:methodology}, the training data set is the set of $n$ days preceding the test set. Hence with increase in training size, the new data added is for months belonging to a different season as compared to our test set, leading to a difference in training and test distribution. This hypothesis will be investigated further in future work. 
% Note that increasing training data seems to have no big impact on the performance of local TFT models. This could be due the training and inference method used~\cref{sec:methodology}. With increase in training size, the new data added corresponds to months from a different season as compared to the test est, leading to a difference in training and test distribution. This hypothesis will be investigated further in future work. 

\subsection{Control Performance}
\begin{figure}[t]
    \centering
    \includegraphics[width=0.95\columnwidth]{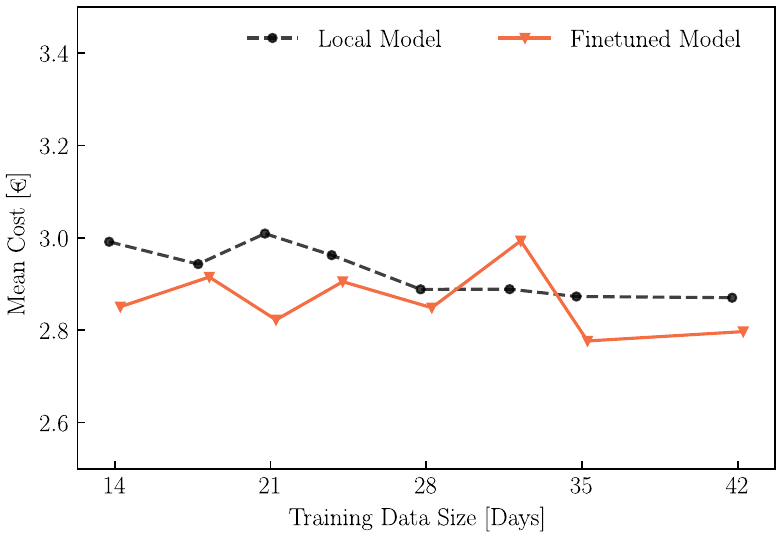}
    \caption{Comparing mean performance of MPC using forecasts from fine-tuned model and a locally trained model over a 1-week period for all 5-test houses.}
    \label{fig:control_perf}
\end{figure}

Following the forecasting performance, we now investigate the impact of the forecasts obtained from our finetuned TFT models on the quality of control policies for a simple HEM system. Based on the MPC presented in~\cref{sec:problem_formulation} and using BELPEX day-ahead electricity prices, we evaluate the performance of a control strategy that uses a small residential battery~(5kW, 10kWh) to reduce the household's energy cost over a 7-day period. \Cref{fig:control_perf} shows the mean performance of such an MPC while using forecasts from either our finetuned or local model over the 5-test houses. It can be observed that the MPC with the finetuned model performs slightly better than the local trained models, with an overall cost reduction of $\sim 2 \%$. However, this difference in performance between the two controllers is lower as compared to the improvement observed for the forecasting performance, with the performance of finetuned models even dropping for a few training sizes. This performance gap can be due to simplistic nature of the MPC, the dimensioning of the battery or the small test size~(1 week) used for these experiments and will be investigated further in future work.

%===================================================================
\section{Conclusion}
\label{sec:conclusions}
%===================================================================

% - Pre-training and TL can help use data effectively 
% - Allows use of SOTA TFT models that are data intensive
% - Provides good performance for individual assets on high freq-short term

% - Future work on improving tl methods -- freezing/unfreezing key layers-- new loss for more household specific loss -- etc. 
% - Future work on implementing improved control -- scenario/stochastic control/ model-based RL 

Based on the results presented in~\cref{sec:results}, it is evident that for low training sizes, the finetuned models perform better than the models trained only on individual household data.
% These results indicate that leveraging data from multiple households to pretrain a global model and then finetuning it for individual households significantly improves demand forecasts. 
This supports our hypothesis that using data from different households in a transfer learning-based approach can lead to data-efficient forecasting models that can produce good quality forecasts and can be used in advanced controllers such as MPCs to develop home energy management systems.
Following up on these results, we plan to expand the scope of our study in two main research areas. The first one focuses on improving the fine-tuning methodology. This involves creating domain-specific finetuning strategies that can leverage the temporal representations learnt by the global model and combine prior knowledge about the household to fine-tune the forecaster more efficiently. The other research direction will focus on open-source contributions. This involves pretraining global models on a larger set of household data and integrating these global models with platforms such as HuggingFace~\cite{hugging_face}.\footnote{This is out-of-scope for the work/models presented in this work due to contractual limitations. } 
% This will enable a wide range of applications and more importantly control architectures, to easily access good quality forecasts and develop home energy managements solutions that can easily scale to new households.

%===================================================================
\section*{Acknowledgements}
%===================================================================
This project was performed in collaboration with Centrica Business Solutions (CENTRICA) and we thank the CENTRICA team for their support throughout this study. 
This work received funding from the European Union's Horizon 2020 research and innovation programme under the projects BD4NRG~(grant no.\ 872613) and BRIGHT~(grant no.\ 957816)

%%
%% The next two lines define the bibliography style to be used, and
%% the bibliography file.

%===================================================================
\bibliographystyle{ACM-Reference-Format}
\bibliography{literature}

%===================================================================

\end{document}